\documentclass{amia}
\usepackage{graphicx}
\usepackage[font=small,labelfont=bf]{caption}
\usepackage[superscript,nomove]{cite}
\usepackage{color}
\usepackage{hyperref}
\usepackage{enumitem}
\usepackage{todonotes}

\usepackage{wrapfig}
\graphicspath{{images/}}
\hypersetup{colorlinks=true,linkcolor=blue,urlcolor=blue}
\begin{document}
\title{Disease Progression Modeling Workbench 360}

\author{
    Parthasarathy Suryanarayanan$^{1}$, 
    Prithwish Chakraborty$^{1}$,
    Piyush Madan$^{1}$,
    Kibichii Bore$^{2}$,
    William Ogallo$^{2}$,
    Rachita Chandra$^{1}$,
    Mohamed Ghalwash$^{1}$,
    Italo Buleje$^{1}$,
    Sekou Remy$^{2}$,
    Shilpa Mahatma$^{1}$,
    Pablo Meyer$^{1}$,
    Jianying Hu$^{1}$
}

\institutes{
    $^1$Center for Computational Health, IBM Research, NY, USA; \\
    $^2$IBM Research, Nairobi, Kenya
}

\maketitle
\section{Background}
Disease Progression Modeling (DPM)~\cite{wang2019survey} 
aims to characterize the progression of a disease and its comorbidities over time using a wide range of analytics models including disease staging~\cite{sun2019probabilistic}
, patient trajectory analytics~\cite{dey2021impact}, prediction~\cite{prithwish2021amia}, and time-to-event estimations~\cite{liu2018early} for key disease-related events. 
DPM has applications throughout the healthcare ecosystem, from providers (e.g., decision support for patient staging), to payers (e.g., care management), and pharmaceutical companies (e.g., clinical trial enrichment). But the complexity of building effective DPM models can be a road-block for their rapid experimentation and adoption. 
Some of this 
is addressed by standardization of data model and tooling for data analysis and cohort selection~\cite{hripcsak2016characterizing}. However, there are still 
unmet needs 
to facilitate the development of advanced machine learning techniques such as deep learning with 
additional requirements such as experiment tracking and reproducibility~\cite{mcdermott2019reproducibility}.
Furthermore, to accelerate DPM research, a data scientist's available tools should include a framework for deploying models as cloud-ready microservices for rapid prototyping and dissemination~\cite{makinen2021needs}.

\setlength\intextsep{-19pt}
\begin{wrapfigure}{r}{0.3\textwidth}
    \centering
    \includegraphics[width=\linewidth]{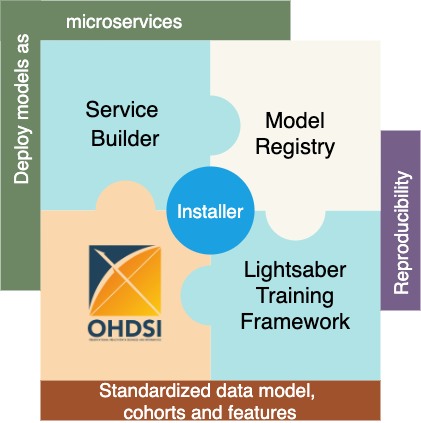}
    \caption{\small DPM360\vspace{1 em} Component View}
    \label{fig:dpm360}
\end{wrapfigure}

In  this  demonstration, we introduce Disease Progression Modeling  Workbench 360 (DPM360) open source project (\href{https://ibm.github.io/DPM360/}{https://ibm.github.io/DPM360/}). DPM360 is an easy-to-install system to help research and development of DPM models (Figure~\ref{fig:dpm360}). 
It manages the entire modeling life cycle, from data analysis (e.g, cohort identification) to machine learning algorithm development and prototyping. DPM360 augments the advantages of data model standardization and tooling (OMOP-CDM, Athena, ATLAS) provided by the widely-adopted OHDSI~\cite{hripcsak2015observational} initiative, with a powerful machine learning training framework, and a mechanism for rapid prototyping through automatic deployment of models as containerized services into a cloud environment. This enables a quicker and flexible implementation and validation of the models.

\section{Methods}

\begin{figure}[h]
    \centering
    \includegraphics[width=0.9\linewidth]{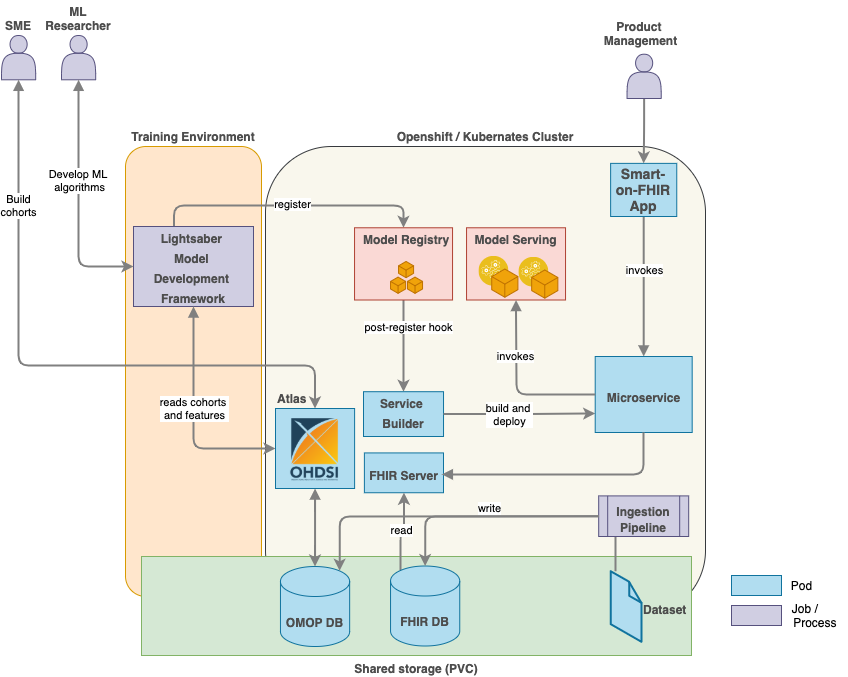}
    \caption{DPM360 Architecture}
    \label{fig:dpm360-full-arch-public}
\end{figure}

The architecture, shown in Figure~\ref{fig:dpm360-full-arch-public} has four main components. 

(1) \textbf{Lightsaber}: an extensible training framework which provides blueprints for the development of disease progression models (DPM). It is designed ground up using state-of-the art open source tools~\cite{scikit-learn,falcon2019pytorch} to provide a simple modular and unified model training framework to support some of the common use cases for DPM.
Lightsaber contains four key modules: 
\begin{itemize}[noitemsep, topsep=0pt]
    \item \textit{data ingestion} modules to support standardized methods of ingesting data
    \item \textit{model trainers} to support standardized model training incorporating best practices 
    \item \textit{metrics} to support pre-built DPM problem specific model evaluation
    \item in-built model tracking and support for post-hoc model evaluation by integrating with a \textit{Model Registry}.
\end{itemize}
Users can select specific modules and integrate them into their modeling workflow. Lightsaber also comes with a reusable library of state-of-the-art machine and deep learning algorithms for DPM (e.g. LSTM~\cite{gers2000learning} for in-hospital mortality predictions). 

Lightsaber integrates naturally with ATLAS using a client called Lightsaber Client for ATLAS (\textit{LCA}), enabling automated extraction of features from the \href{https://www.ohdsi.org/data-standardization/the-common-data-model/}{OMOP CDM} model, thus complementing the ease and flexibility of defining standardized cohorts using ATLAS graphical user interface with the ability to quickly develop deep learning algorithms for DPM in Lightsaber using Python. \textit{LCA} can be configured with the cohort details, covariate settings, model training settings for Lightsaber to extract the right set of features in formats currently supported in the OHDSI stack (see \href{https://github.com/OHDSI/FeatureExtraction}{FeatureExtraction} and \href{https://github.com/OHDSI/PatientLevelPrediction}{PatientLevelPrediction} R packages via the \href{https://pypi.org/project/rpy2/}{Rpy2} interface). Additionally, the \textit{LCA} uses custom queries and algorithms to extract and transform complex time series features into formats required for DPM in Lightsaber. For each feature extraction process, a YAML configuration file is automatically generated. This file specifies outcomes, covariate types, and file locations of the extracted feature files. Thus, Lightsaber allows a user to concentrate just on the logic of their model as it takes care of the rest.

(2) Tracking provenance of all aspects of model building is essential for trust and reproducibility - thus experiments ran using Lightsaber are automatically tracked in a \textbf{Model Registry} including model parameters, problem specific metrics, and model binaries allowing the identification of algorithms and parameters that result in the best model performance.

(3) The \textbf{Service Builder} component automatically converts registered models in \textit{Model Registry} into microservices~\cite{dragoni2017microservices}, through the usage of hooks that listen for production ready models in the registry and thereafter start the model packaging execution pipeline. The pipeline includes extraction of model and its dependencies from the registry, containerization, and deployment in the target cluster (\href{https://kubernetes.io}{Kubernates} or \href{https://www.openshift.com}{OpenShift}). Upon successful model deployment, a callback function updates model metadata in the registry with deployment status and model access endpoint. Using this endpoint, potential users (data scientist or product manager) can interact with the model, now deployed as a microservice, though a \href{https://swagger.io/}{Swagger} based interface.

(4) The \textbf{Installer} component installs the fully functional DPM360, including OHDSI tools, \textit{Model Registry}, and \textit{Service Builder} into a Kubernetes or OpenShift cluster using \href{https://helm.sh/}{Helm charts}. Each of these components are run as services within the cluster. The implementation also uses \href{https://kubernetes.io/docs/concepts/storage/persistent-volumes/}{Persistent Volume Claims} to persist the data (e.g: model artifacts, ATLAS database files etc.).

\section{Results}
A detailed description of the target use cases, architecture and road map is available in our public GitHub repository at \url{https://ibm.github.io/DPM360/} which contains our initial implementation. To verify the integration of the overall system, we reproduced the results from clinical prediction benchmarks from MIMIC III~\cite{Harutyunyan}. Specifically, we defined and extracted cohorts related to in-hospital mortality in critical care settings from an OMOP CDM version of the MIMIC III dataset~\cite{johnson2016mimic}. Using ATLAS, we defined our target cohort consisting of adult patients who have been hospitalized for the first time for at least two days, and who have at least one measurement recorded within the first 48 hours, and our outcome cohort of adult patients who died in hospital within 30 days of the first admission. We subsequently used the LCA to extract the features for the prediction task, and then trained a model using the Lightsaber framework. We obtained comparable metrics to the benchmark, thus proving the soundness of the overall system. Also, independently, the Lightsaber component has been successfully used for rapid experimentation and analysis of real-world data~\cite{prithwish2021amia,dey2021impact}.

\section{Conclusion}
In the coming months, we plan to execute our published road-map for DPM360 and report on its effectiveness at improving collaborative research and rapid commercialization  from several ongoing DPM collaborations.

\section*{Acknowledgements}
We would like to thank Divya Pathak and Daby Sow for their guidance, and Sundar Saranthan for his contributions.

\makeatletter
\renewcommand{\@biblabel}[1]{\hfill #1.}
\makeatother

\setlength\itemsep{-0.1em}

\end{document}